\documentclass{article}

\usepackage{arxiv}

\usepackage[utf8]{inputenc} 
\usepackage[T1]{fontenc}    
\usepackage{url}            
\usepackage{booktabs}       
\usepackage{amsfonts}       
\usepackage{nicefrac}       
\usepackage{microtype}      
\usepackage{lipsum}
\usepackage{graphicx}

\usepackage{amsmath}
\usepackage{amsthm}
\usepackage{amssymb}
\usepackage{mathtools}
\usepackage{arydshln}
\RequirePackage{algorithm}

\usepackage[accsupp]{axessibility}  

\usepackage[pagebackref,breaklinks,colorlinks]{hyperref}

\usepackage[capitalize]{cleveref}
\crefname{section}{Sec.}{Secs.}
\Crefname{section}{Section}{Sections}
\Crefname{table}{Table}{Tables}
\crefname{table}{Tab.}{Tabs.}

\usepackage{amsthm}
\usepackage{svg}
\svgsetup{inkscapeformat=png}
\usepackage{soul}
\usepackage{mathtools}
\usepackage{hyperref}
\usepackage{amssymb}
\usepackage{bm}
\usepackage{float}
\usepackage{amsmath}
\usepackage{tikz}
\usetikzlibrary{calc}
\usepackage{bbm}
\usetikzlibrary{positioning, backgrounds, fit, shapes.arrows}
\usetikzlibrary{decorations.markings}
\usepackage{cuted}

\usepackage{enumitem}
\usepackage{caption}
\usepackage{subcaption}
\usepackage{booktabs} 
\usepackage{adjustbox} 
\usepackage{multirow} 

\usepackage{algorithm}
\usepackage{algpseudocode}
\tikzset{block/.style = {draw, fill=white, rectangle,
		minimum height=3em, minimum width=2cm},
	input/.style = {coordinate},
	output/.style = {coordinate},
	pinstyle/.style = {pin edge={to-,t,black}}
	radiation/.style={{decorate,decoration={expanding waves,angle=90,segment   length=4pt}}}
	
}
\usepackage{smartdiagram}

\tikzstyle{block} = [draw, rectangle, minimum height=2em, minimum width=2em]
\tikzstyle{sum} = [draw, circle,minimum width=0.1 cm]
\tikzstyle{input} = [coordinate]
\tikzstyle{output} = [coordinate]
\tikzstyle{dummy} = [coordinate]
\tikzstyle{pinstyle} = [pin edge={to-,thin,black}]
\usetikzlibrary{positioning, fit, arrows.meta}
\usetikzlibrary{positioning}
\usetikzlibrary{shapes,arrows}
\tikzstyle{frame_cyan} = [thick, draw=blue, solid,inner sep=0.3em]
\tikzstyle{frame_red} = [thick, draw=red, solid,inner sep=0.3em]
\tikzstyle{frame_green} = [thick, draw=green, solid,inner sep=0.3em]

\usepackage{tikz}
\usepackage{xcolor}
\definecolor{fc}{HTML}{1E90FF}
\definecolor{h}{HTML}{228B22}
\definecolor{bias}{HTML}{87CEFA}
\definecolor{noise}{HTML}{8B008B}
\definecolor{conv}{HTML}{FFA500}
\definecolor{pool}{HTML}{B22222}
\definecolor{up}{HTML}{B22222}
\definecolor{view}{HTML}{FFFFFF}
\definecolor{bn}{HTML}{FFD700}
\tikzset{fc/.style={black,draw=black,fill=fc,rectangle,minimum height=1cm}}
\tikzset{h/.style={black,draw=black,fill=h,rectangle,minimum height=1cm}}
\tikzset{bias/.style={black,draw=black,fill=bias,rectangle,minimum height=1cm}}
\tikzset{noise/.style={black,draw=black,fill=noise,rectangle,minimum height=1cm}}
\tikzset{conv/.style={black,draw=black,fill=conv,rectangle,minimum height=1cm}}
\tikzset{pool/.style={black,draw=black,fill=pool,rectangle,minimum height=1cm}}
\tikzset{up/.style={black,draw=black,fill=up,rectangle,minimum height=1cm}}
\tikzset{view/.style={black,draw=black,fill=view,rectangle,minimum height=1cm}}
\tikzset{bn/.style={black,draw=black,fill=bn,rectangle,minimum height=1cm}}
 
\usepackage{xspace}

\tikzstyle{dummy} = [coordinate]
\pgfkeys{/pgf/.cd,
  parallelepiped offset x/.initial=2mm,
  parallelepiped offset y/.initial=2mm
}
\pgfdeclareshape{parallelepiped}
{
  \inheritsavedanchors[from=rectangle] 
  \inheritanchorborder[from=rectangle]
  \inheritanchor[from=rectangle]{north}
  \inheritanchor[from=rectangle]{north west}
  \inheritanchor[from=rectangle]{north east}
  \inheritanchor[from=rectangle]{center}
  \inheritanchor[from=rectangle]{west}
  \inheritanchor[from=rectangle]{east}
  \inheritanchor[from=rectangle]{mid}
  \inheritanchor[from=rectangle]{mid west}
  \inheritanchor[from=rectangle]{mid east}
  \inheritanchor[from=rectangle]{base}
  \inheritanchor[from=rectangle]{base west}
  \inheritanchor[from=rectangle]{base east}
  \inheritanchor[from=rectangle]{south}
  \inheritanchor[from=rectangle]{south west}
  \inheritanchor[from=rectangle]{south east}
  \backgroundpath{
    \southwest \pgf@xa=\pgf@x \pgf@ya=\pgf@y
    \northeast \pgf@xb=\pgf@x \pgf@yb=\pgf@y
    \pgfmathsetlength\pgfutil@tempdima{\pgfkeysvalueof{/pgf/parallelepiped offset x}}
    \pgfmathsetlength\pgfutil@tempdimb{\pgfkeysvalueof{/pgf/parallelepiped offset y}}
    \def\ppd@offset{\pgfpoint{\pgfutil@tempdima}{\pgfutil@tempdimb}}
    \pgfpathmoveto{\pgfqpoint{\pgf@xa}{\pgf@ya}}
    \pgfpathlineto{\pgfqpoint{\pgf@xb}{\pgf@ya}}
    \pgfpathlineto{\pgfqpoint{\pgf@xb}{\pgf@yb}}
    \pgfpathlineto{\pgfqpoint{\pgf@xa}{\pgf@yb}}
    \pgfpathclose
    \pgfpathmoveto{\pgfqpoint{\pgf@xb}{\pgf@ya}}
    \pgfpathlineto{\pgfpointadd{\pgfpoint{\pgf@xb}{\pgf@ya}}{\ppd@offset}}
    \pgfpathlineto{\pgfpointadd{\pgfpoint{\pgf@xb}{\pgf@yb}}{\ppd@offset}}
    \pgfpathlineto{\pgfpointadd{\pgfpoint{\pgf@xa}{\pgf@yb}}{\ppd@offset}}
    \pgfpathlineto{\pgfqpoint{\pgf@xa}{\pgf@yb}}
    \pgfpathmoveto{\pgfqpoint{\pgf@xb}{\pgf@yb}}
    \pgfpathlineto{\pgfpointadd{\pgfpoint{\pgf@xb}{\pgf@yb}}{\ppd@offset}}
  }
}
\pgfdeclareshape{document}{
\inheritsavedanchors[from=rectangle] 
\inheritanchorborder[from=rectangle]
\inheritanchor[from=rectangle]{center}
\inheritanchor[from=rectangle]{north}
\inheritanchor[from=rectangle]{north east}
\inheritanchor[from=rectangle]{north west}
\inheritanchor[from=rectangle]{south}
\inheritanchor[from=rectangle]{south east}
\inheritanchor[from=rectangle]{south west}
\inheritanchor[from=rectangle]{west}
\inheritanchor[from=rectangle]{east}
\backgroundpath{%
\southwest \pgf@xa=\pgf@x \pgf@ya=\pgf@y
\northeast \pgf@xb=\pgf@x \pgf@yb=\pgf@y
\pgf@xc=\pgf@xb \advance\pgf@xc by-5pt 
\pgf@yc=\pgf@ya \advance\pgf@yc by5pt
\pgfpathmoveto{\pgfpoint{\pgf@xa}{\pgf@ya}}
\pgfpathlineto{\pgfpoint{\pgf@xa}{\pgf@yb}}
\pgfpathlineto{\pgfpoint{\pgf@xb}{\pgf@yb}}
\pgfpathlineto{\pgfpoint{\pgf@xb}{\pgf@yc}}
\pgfpathlineto{\pgfpoint{\pgf@xc}{\pgf@ya}}
\pgfpathclose
\pgfpathmoveto{\pgfpoint{\pgf@xc}{\pgf@ya}}
\pgfpathlineto{\pgfpoint{\pgf@xc}{\pgf@yc}}
\pgfpathlineto{\pgfpoint{\pgf@xb}{\pgf@yc}}
\pgfpathclose
}
}
\tikzstyle{block} = [draw, fill=white, rectangle, 
    minimum height=3em, minimum width=6em]
    
\usetikzlibrary{backgrounds}
\usepackage{pifont}
\tikzstyle{startstop} = [rectangle, rounded corners, minimum width=2cm, minimum height=0.7cm,text centered, draw=black, fill=lime!30]
\usepackage[textsize=scriptsize,textwidth=1.4cm, disable]{todonotes}
\usetikzlibrary{matrix}
\usepackage{xcolor,colortbl}

\newcommand\MakeUppercaseGreek[1]{
  \begingroup
    \let\psi\Psi
    \let\omega\Omega
    \let\gamma\Gamma
    \MakeUppercase{#1}
  \endgroup}
\newcommand{\vectorsym}[1]{\bm{#1}}
\newcommand{\expectation}[2]{\mathbb{E}_{#2}\left[#1\right]}

\newcommand{\brackets}[1]{\left(#1\right)}
\newcommand{\abs}[1]{\left|#1\right|}
\newcommand{\norm}[1]{\left\lVert#1\right\rVert}

\newcommand{\squareb}[1]{\left[{#1}\right]}

\newcommand{\cmark}{\ding{51}}%
\newcommand{\tstd}[1]{{\color{gray}\tiny $\pm#1$}}

 \newcommand{\mul}[0]{\vectorsym{\mu}_l}
\newcommand{\sigmal}[0]{\vectorsym{\sigma}_l}
\newcommand{\mubar}[0]{\bar{\vectorsym{\mu}}_l}
\newcommand{\sigmabar}[0]{\bar{\vectorsym{\sigma}}_l}

\newcommand{\xset}[0]{\bm{X}}
\newcommand{\yset}[0]{\bm{Y}}
\newcommand{\bset}[0]{\bm{B}}
\newcommand{\dset}[0]{\bm{\mathcal{D}}}

\newcommand{\faug}[0]{g_{aug}}

\newcommand{\secondbest}[1]{\textbf{#1}}
\newcommand{\best}[1]{\colorbox{black!30}{\textbf{#1}}}
\newcommand{\zeroq}[0]{ZeroQ\cite{cai2020_zeroq}}
\newcommand{\intraq}[0]{IntraQ\cite{Zhong2022_IntraQ}}
\newcommand{\genie}[0]{GENIE\cite{jeon2023genie}}
\newcommand{\name}[0]{DGH}

\newcommand{\x}[0]{\vectorsym{x}}

\usepackage{caption}
\begin{document}
\definecolor{MyGreen}{rgb}{0.0, 0.7, 0.0}

\title{Data Generation for Hardware-Friendly Post-Training Quantization} 

\author{
 Lior Dikstein \\
  Sony Semiconductor Israel\\
  \texttt{lior.dikstein@sony.com} \\
   \And
 Ariel Lapid \\
  Sony Semiconductor Israel\\
  \texttt{ariel.lapid@sony.com}
    \And
 Arnon Netzer \\
  Sony Semiconductor Israel\\
  \texttt{arnon.netzer@sony.com}  
  \And
 Hai Victor Habi \\
  Sony Semiconductor Israel\\
  \texttt{hai.habi@sony.com}
}

\maketitle
\begin{abstract}
    Zero-shot quantization (ZSQ) using synthetic data is a key approach for post-training quantization (PTQ) under privacy and security constraints. However, existing data generation methods often struggle to effectively generate data suitable for hardware-friendly quantization, where all model layers are quantized. We analyze existing data generation methods based on batch normalization (BN) matching and identify several gaps between synthetic and real data: 1) Current generation algorithms do not optimize the entire synthetic dataset simultaneously; 2) Data augmentations applied during training are often overlooked; and 3) A distribution shift occurs in the final model layers due to the absence of BN in those layers. These gaps negatively impact ZSQ performance, particularly in hardware-friendly quantization scenarios.  In this work, we propose Data Generation for Hardware-friendly quantization (\name{}), a novel method that addresses these gaps. \name{} jointly optimizes all generated images, regardless of the image set size or GPU memory constraints. To address data augmentation mismatches, \name{} includes a preprocessing stage that mimics the augmentation process and enhances image quality by incorporating natural image priors. Finally, we propose a new distribution-stretching loss that aligns the support of the feature map distribution between real and synthetic data. This loss is applied to the model's output and can be adapted to various tasks. \name{} demonstrates significant improvements in quantization performance across multiple tasks, achieving up to a 30$\%$ increase in accuracy for hardware-friendly ZSQ in both classification and object detection, often performing on par with real data.  
\end{abstract}
\section{Introduction}
\label{sec:intro}

\par Deep learning models (DNNs) are revolutionizing the field of computer vision, but deploying them on devices with limited memory or processing power remains a challenge \cite{cheng2017survey,han2015deep,li2019edge,hubara2018quantized,wang2019haq}. Quantization, a powerful optimization technique, compresses the size of the model and reduces the computational demands, enabling efficient deployment \cite{krishnamoorthi2018quantizing,nagel2021white}.
While quantization-aware training (QAT) incorporates quantization during the training process to help maintain the original accuracy  \cite{jacob2018quantization, esser2020learned}, post-training quantization (PTQ)\cite{habi2021hptq,banner2019_post, choukroun2019low,nagel2020_adaround,li2021_brecq} has gained significant traction for its ability to compress the model without requiring retraining. However, PTQ's effectiveness depends on the availability of representative data that aligns with the statistics of the original training set \cite{hubara2021accurate,zhong2022fine,xu2020generative, liu2023pd, zheng2022leveraging, zhang2018lq, zhou2017incremental}.
 
\par In many practical scenarios, access to such representative data samples is limited by privacy or security concerns. Regulatory restrictions or proprietary constraints may prevent the collection or use of even a few samples during model deployment. Zero-shot quantization (ZSQ) offers an alternative solution, enabling PTQ by utilizing synthetic data generation.

State-of-the-art data generation methods often rely on batch normalization (BN) \cite{ioffe2015batch} statistics matching to ensure the generated data aligns with the real data distribution \cite{cai2020_zeroq,Zhang2021_DSG,zhong2022fine,xu2020generative,yin2020dreaming,haroush2019,Zhong2022_IntraQ,jeon2023genie}.  Despite considerable advancements, these methods often fail to generate data that is suitable for hardware-friendly quantization, where all model layers are quantized. To overcome this challenge, we analyze current methods and identify three key gaps in the generation process. The first gap lies in the inconsistency between how BN statistics are collected during training and how current data generation methods handle them. Typically, BN aggregates statistics across the entire dataset. However, existing data generation techniques optimize each sample \cite{cai2020_zeroq} or batch \cite{Zhong2022_IntraQ,Zhang2021_DSG,haroush2019, yin2020dreaming,jeon2023genie} independently. This can result in synthetic datasets that fail to accurately represent the global statistical properties and diversity of the training data. The second gap stems from the data augmentation applied during training, which affects BN-collected statistics. These effects are often overlooked in synthetic data generation. The final gap occurs since existing methods focus primarily on minimizing a loss which is derived from BN statistics. However, there is no BN comparison point towards the later stages of deep neural networks, particularly at the output. This results in a distribution mismatch between the feature maps of generated data and real data at these stages, a critical factor for hardware-friendly quantization schemes \cite{habi2020hmq,li2021mqbench,habi2021hptq} that quantize the entire model, especially the output layer.

We address these gaps and introduce a Data Generation method for Hardware-friendly post-training quantization (\name{}) that enables ZSQ on real-world devices. To handle the BN statistic collection gap, \name{} generates images while computing the BN statistics over the entire set of generated images. This is achieved by using a statistical aggregation approach that is not limited by available GPU memory. Moreover, optimizing BN statistics over the entire set relaxes the constraints on individual images, allowing them to deviate from the global mean and standard deviation, better approximating real data characteristics. Additionally, to account for the effects of data augmentation and to incorporate characteristics typical of natural images, we introduce an image-enhancement preprocessing step that includes smoothing operations and data augmentation. Finally, to address the stages unaffected by the BN loss, we propose a novel output distribution stretching loss mechanism that encourages generated images to utilize the model output's dynamic range. This loss can be applied to any task. This loss combined with the statistical aggregation scope are critical for generating images for ZSQ in deployment environments requiring full model quantization.

\par To summarize, our contributions are as follows:
\begin{itemize}
    \item We identify and analyze three key gaps in existing data generation methods: the optimization scope, the impact of overlooking data augmentation, and the distribution mismatch in later network stages where BN layers are absent.
    \item We present \name{}, a data generation method that enables hardware-friendly ZSQ. The method integrates three key components: statistics aggregation across the entire set of generated images, image enhancement preprocessing, and an output distribution stretching loss.
    \item We present experimental results on both ImageNet-1k classification (ResNet18, ResNet50 and MobileNet-V2) and COCO object detection (RetinaNet, SSDLiteMobilenetV3, and FCOS), demonstrating \name's benefits for hardware-friendly ZSQ. For instance, \name $ $ achieves up to 30$\%$ improvement over all state-of-the-art methods in hardware-friendly quantization. Additionally, we provide an ablation study that analyzes \name{} and highlights the impact of addressing all identified gaps.
\end{itemize}

In the spirit of reproducible research, the code of this work is available at: \href{https://github.com/ssi-research/DGH}{https://github.com/ssi-research/DGH}. The algorithm is also available as part of Sony's open-source Model-Compression-Toolkit library at: \href{https://github.com/sony/model_optimization}{https://github.com/sony/model\_optimization}.

\section{Related Work}
\subsection{Quantization}
\par Quantization methods are broadly divided into two categories \cite{gholami2022survey, liang2021pruning, nagel2021white}. The first category is QAT\cite{nagel2022overcoming, fan2020training, esser2020learned, jacob2018quantization, hubara2018quantized, courbariaux2015binaryconnect, bhalgat2020lsq+}, which fine-tunes model weights by alternately performing quantization and backpropagation on abundant labeled training data. The performance achieved by QAT is often comparable to that of the respective floating-point models. However, this strategy is impractical in many real-world scenarios due to the expensive and time-consuming computation, algorithmic complexity, and the need for access to the entire training dataset. 
\par The second category is PTQ \cite{habi2021hptq,nagel2020_adaround,hubara2021accurate,li2021_brecq,wei2022qdrop, nahshan2021loss, jeon2022mr}, offering advantages in terms of data requirements and computational efficiency compared to QAT. Although PTQ methods do not rely on the availability of complete labeled training data, they require a small set of unlabeled domain samples for calibration.

\subsection{Data Generation}

\par Zero-shot quantization (ZSQ) methods aim to eliminate the need for representative data during quantization \cite{choi2020data}. These methods can be roughly divided into two categories. The first category includes methods that operate without any data, such as DFQ \cite{nagel2019_dfq}, Squant \cite{guo2022_squant}, and OCS \cite{zhao2019improving}. However, these methods are unsuitable for low-bit precision or activation quantization.

\par The second category leverages synthetic data generation for calibration. These methods exploit the inherent knowledge within a pre-trained model to generate synthetic images that closely resemble the true data distribution. One approach within this category is generator-based, where a generator is trained alongside the quantized model \cite{xu2020generative, choi2021qimera, liu2021zero, zhu2021autorecon, choi2022s, li2024acq}. The generator is trained to produce synthetic images with properties similar to the training data. However, generator-based methods demand extensive computational resources, as the generator requires training from scratch for each bit-width configuration. 

\par Another approach treats data synthesis as an optimization problem. These methods are generally much faster than generator-based approaches. Methods such as ZeroQ\cite{cai2020_zeroq}, DeepInversion \cite{yin2020dreaming} and Harush \etal \cite{haroush2019} begin with random noise images and iteratively update them by extracting the mean and standard deviation of activations from the BN layers, aligning them with the model's BN parameters. DSG \cite{Zhang2021_DSG} revealed that distilled images optimized to match BN statistics alone suffer from homogenization compared to real data. To address this, the authors suggested relaxing the BN statistics alignment constraint by introducing margin constants per layer and proposed two additional objectives to diversify the distilled data at the sample level. IntraQ \cite{Zhong2022_IntraQ} further demonstrated that enhancing inter-class and intra-class heterogeneity in the distilled data used for calibration can improve the performance of a quantized model. 
\par GENIE \cite{jeon2023genie} leverages both approaches by using random inputs as latent vectors, passing them through a generator, and feeding the outputs into the model. The generator parameters, along with the random inputs, are updated iteratively. While this method delivers superior quantization results, it is considerably slower due to the need to train the generator from scratch for each batch.
\section{Notation and Preliminaries}
\par To generate a synthetic dataset, $\dset$, for use in quantization, previous methods \cite{cai2020_zeroq, Zhong2022_IntraQ, haroush2019, yin2020dreaming, jeon2023genie} typically utilize statistics, namely, the mean and standard deviation, extracted from the BN layers of pre-trained models. Specifically, let $f_l$ {$:\mathcal{I}\rightarrow \mathbb{R}^{c_l \times d_l}$} denotes the output of the $l^{th}$ layer, where $\mathcal{I}$ represents the input domain, $c_l$ refers to the number of channels in layer $l$ and $d_l$ represents the feature dimensions in $l^{th}$ layer. For example, in the case of image data, $d_l$ represents the spatial dimensions $ d_l=  h_l  w_l$. Next $\faug:\mathcal{I}\rightarrow\mathcal{I}$ denotes the augmentation function, then the BN statistics of the $l^{th}$ layer are denoted as $\mul\in\mathbb{R}^{c_l}$ and $\sigmal\in\mathbb{R}^{c_l}$ given by:

\begin{subequations}
\label{eq:bn_orig_stats}
\begin{align}
    \squareb{\mul}_i = \expectation{\frac{1}{d_l}\sum_{j}\squareb{\yset_l}_{i,j}}{\yset_l}, 
\end{align}

\begin{align}
    \squareb{ \sigmal}_i=\expectation{\frac{1}{d_l}\sum_{j}\brackets{\squareb{\yset_l}_{i,j}-\squareb{\mul}_i}^2}{\yset_l},
\end{align}
\end{subequations}

where $\yset_l = f_l\brackets{\faug\brackets{\xset}}$ is the output of the $l^{th}$ layer during training and $\xset$ represents the distribution of the training set. Given a batch of $K$ images $\bset \triangleq \{\x_k \}_{k=1}^K$ to generate, the BN statistics (BNS) loss is defined as:
\begin{align}
\label{eq:bn_loss}
\mathcal{L}_{BNS}(\bset) =\sum_{l=1}^L&\| \mubar(\bset)-\mul\|_2^2 + \| \sigmabar(\bset)-\sigmal \|_2^2,
\end{align}
where $L$ is the number of BN layers, $\norm{\vectorsym{a}}_2\triangleq\sqrt{\sum_{i=1}\squareb{\vectorsym{a}}_i^2}$ denotes the L2 norm of vector $\vectorsym{a}$, $\squareb{\vectorsym{a}}_i$ denotes the $i^{th}$ element of vector $\vectorsym{a}$, {$\mubar(\bset)$ and $\sigmabar(\bset)$ denote the empirical mean and variance of the $l^{th}$ layer computed over the batch $\bset$. The  empirical mean and variance of the $l^{th}$ layer computed over the batch $\bset$ are given by:}
\begin{subequations}
    \label{eq:activation_stats}
    \begin{align}
    \label{eq:activation_stats_mean}
    \squareb{\mubar(\bset)}_i &= \frac{1}{K d_l}\sum_{k,j=1}^{K,d_l}\squareb{\vectorsym{y}_l^{(k)}}_{i,j},
    \end{align}
    \begin{align}
    \label{eq:activation_stats_var}
    \squareb{\sigmabar(\bset)}_i &= \frac{1}{K d_l}\sum_{k,j=1}^{K,d_l}\brackets{\squareb{\vectorsym{y}_l^{(k)}}_{i,j} - \squareb{\mubar(\bset)}_i}^2,
    \end{align}
\end{subequations}

where $\vectorsym{y}_l^{(k)}=f_l\brackets{\x_k}$ is the feature map of the $l^{th}$ layer and $k^{th}$ sample. State-of-the-art methods, such as \cite{cai2020_zeroq, Zhong2022_IntraQ, haroush2019, yin2020dreaming, jeon2023genie}, divide $\dset = \bigcup_{n=1}^N \bset_n$ into $N$ batches. The loss in \cref{eq:bn_loss} is optimized independently for each batch $\bset_n \in \dset$, where each batch contains $K$ samples, resulting in a dataset of size $\abs{\mathcal{D}}=K\cdot N$. For example, in ZeroQ \cite{cai2020_zeroq}, the statistics of each image are optimized to match those of the entire training set, \ie, $K=1$. In contrast, in \cite{Zhong2022_IntraQ, haroush2019, yin2020dreaming, jeon2023genie}, the statistics for each batch are optimized individually to match the training set, where $K \geq 1$ is a hyperparameter restricted by available GPU memory.

\section{Analyzing Batch Normalization Based Data Generation}
We identified three key gaps between how existing BN-based data generation methods generate data and how models collect statistics during training:
\begin{itemize}
    \item \textbf{Statistics Aggregation Scope:} BN layers aggregate statistics over the entire dataset, while data generation techniques optimize individual batches or images independently to fit the BN statistics.
    
    \item \textbf{Data Augmentation:} Data augmentations applied during training, which affect BN statistics, are often neglected in data generation.
    
    \item \textbf{Output Distribution Mismatch:} The BNS loss has a limited impact on specific feature maps within the model, particularly the model outputs. This discrepancy results in a distribution mismatch between the synthesized and the real data in these feature maps.
    
\end{itemize}

\subsection{Statistics Aggregation Scope}
\begin{figure}[t]
        \centering
        \includegraphics[width=0.8\textwidth]{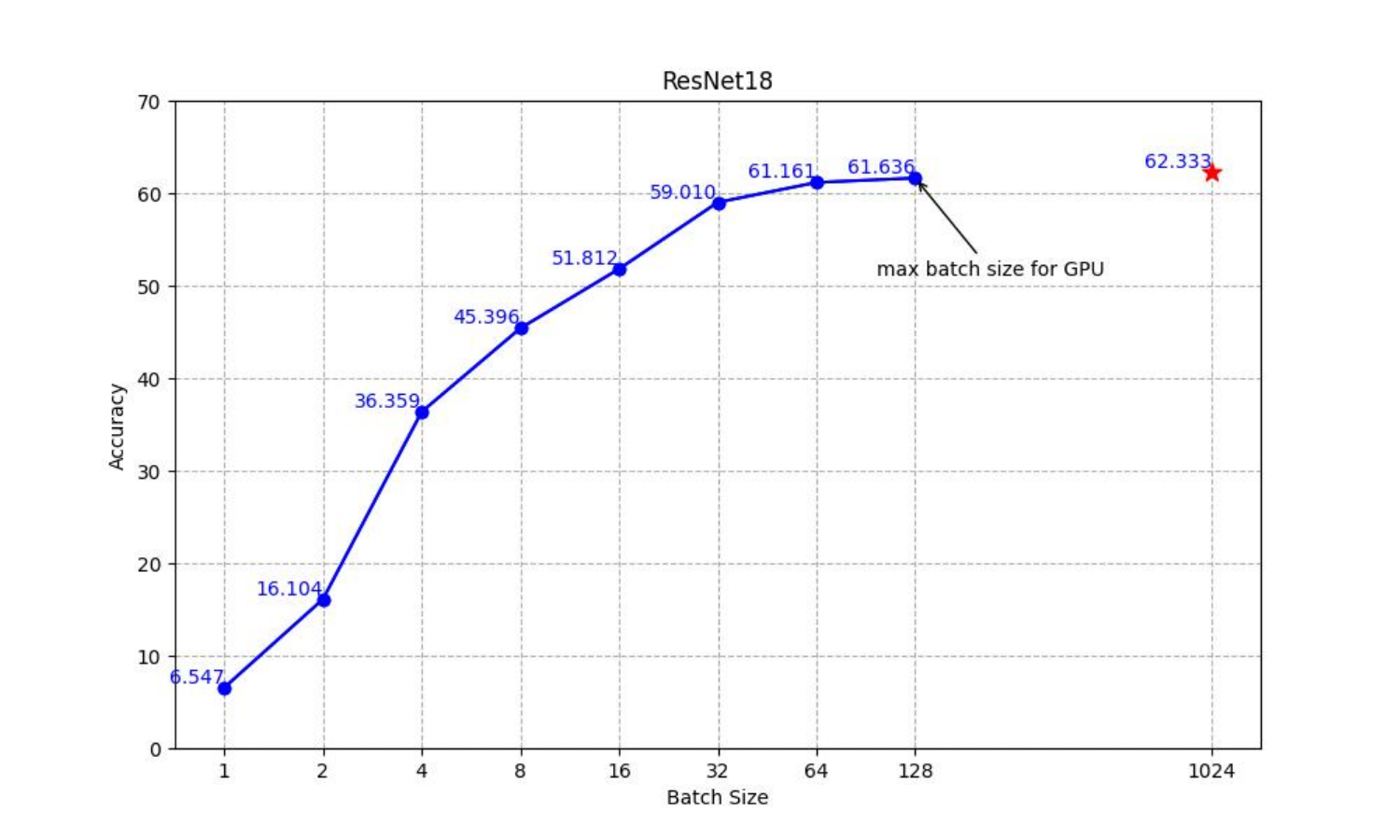}
        \caption{The blue curve represents the Top-1 accuracy on the ImageNet-1k validation set of ResNet-18 quantized to W3A8-bit precision using AdaRound \cite{nagel2020_adaround} with $1024$ generated images, where each batch is optimized separately according to \cref{eq:bn_loss}. The x-axis denotes the statistics aggregation scope (batch size), $K$, used in the image generation process. The red star indicates the result of our proposed aggregation algorithm, which utilizes the statistics of all images collectively.}
    \label{fig:batch_size_effect}
\end{figure}

\par Notably, BN layers calculate mean and standard deviation across the entire dataset, \ie, \cref{eq:bn_orig_stats}, rather than on individual images or batches \cite{ioffe2015batch, bjorck2018understanding}. As a result, a BN layer's statistics represent the mean/std of activations averaged across all training images.
However, existing data generation techniques typically optimize each batch or image individually to match the statistics of the BN layers, \ie, \cref{eq:activation_stats}, ignoring the collective characteristics of the training data.

The impact of the generated images with varying statistics aggregation scopes on the accuracy of ZSQ for a ResNet18 model is illustrated in \cref{fig:batch_size_effect}. 
The curve clearly shows that quantization accuracy varies with the statistics aggregation scope, indicating that smaller scopes (batches) lead to generated images resulting in lower accuracy. In contrast, the red star highlights the accuracy achieved by our proposed aggregation algorithm, which optimizes the statistics across all images simultaneously, resulting in improved performance.
\par
\begin{figure}[ht]
\begin{minipage}[t]{\textwidth}
    \centering
    \includegraphics[width=\textwidth]{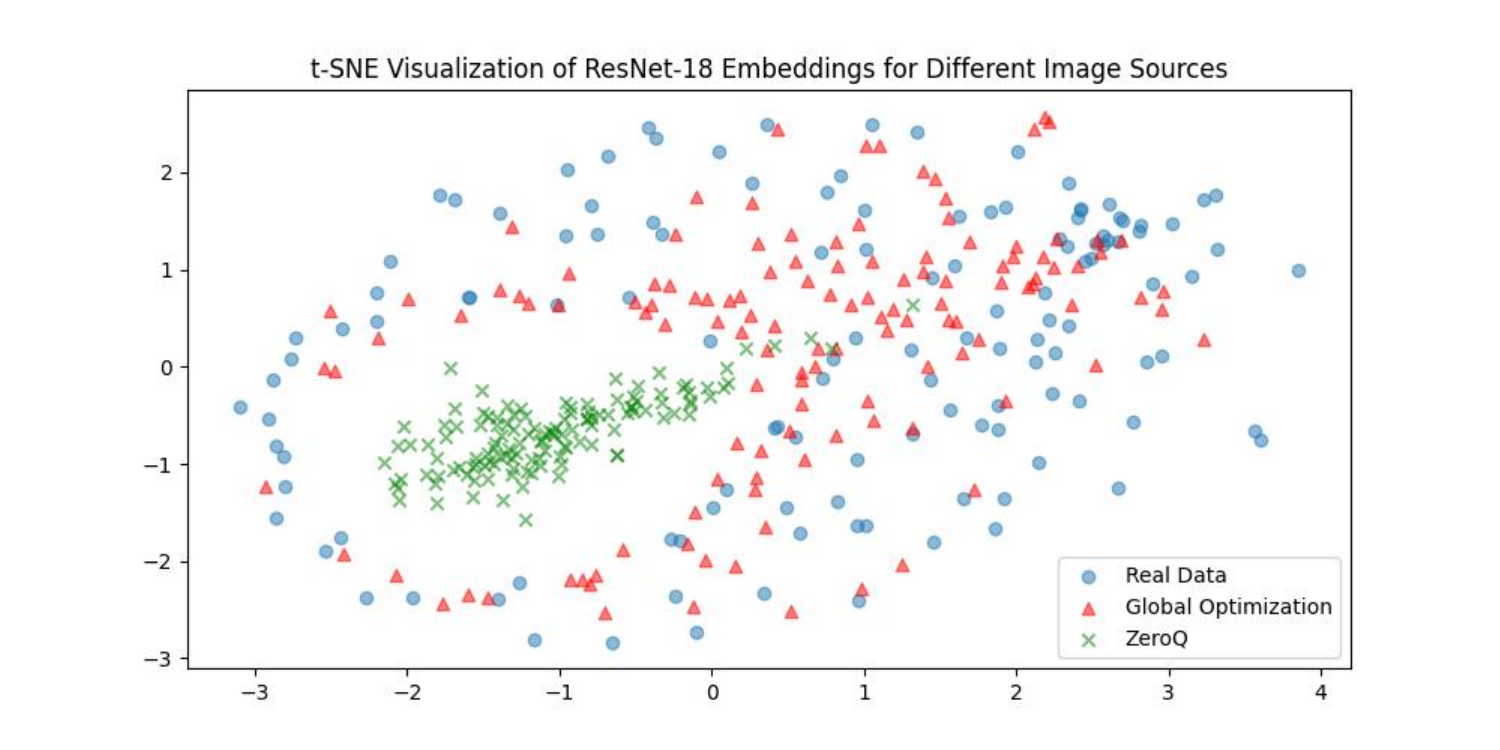}
    \hfill
    \caption{Two-dimensional t-SNE \cite{van2008visualizing} visualization of ResNet-18 embeddings comparing real images (blue) with those generated with global optimization (red) and ZeroQ (green).}
    \label{fig:tsne_real_vs_zeroq_vs_ours}
\end{minipage}
\end{figure}

 To further explore the impact of statistical calculation granularity on image generation, we visualized the embeddings of a ResNet-18 model using t-SNE \cite{van2008visualizing}. \cref{fig:tsne_real_vs_zeroq_vs_ours} shows a two-dimensional t-SNE visualization: comparing real images versus those generated with global optimization and ZeroQ (sample optimization). Notably, embeddings from global optimization substantially overlap with real images, indicating a close match in the embedding space. In contrast, embeddings from ZeroQ are distinct from real images, showing significant separation. This highlights the benefits of our method's global statistic calculation over ZeroQ's per-image approach, suggesting our method more accurately captures the training data distribution.

\subsection{Data Augmentation}
\begin{figure}[ht]
\begin{minipage}[t]{\textwidth}
    \centering
        \centering
        \includegraphics[width=\textwidth]{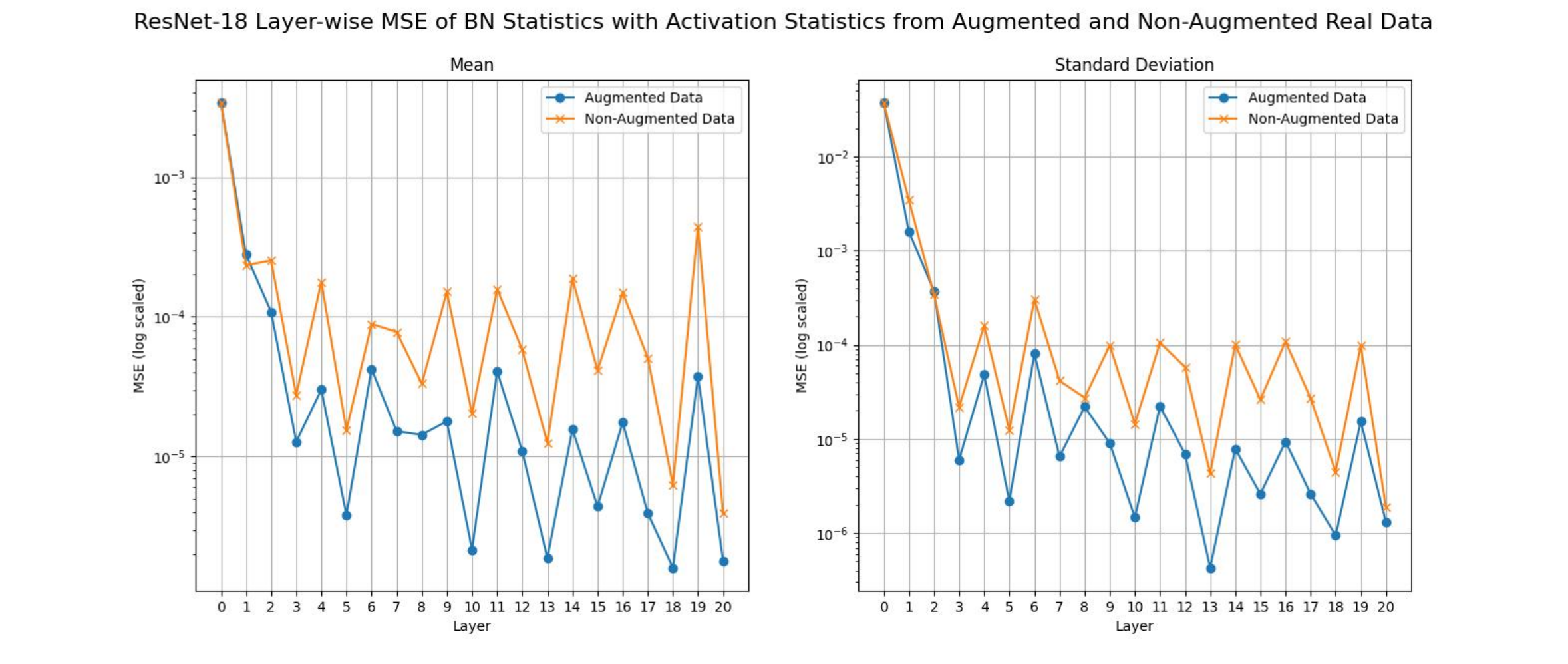}
    \hfill
    \caption{A layer-wise comparison of the MSE for mean (left) and standard deviation (right) of model activations relative to BN statistics in a ResNet-18 model. The blue line indicates MSE values calculated using augmented real data, while the orange line represents values from non-augmented real data. Each variant runs 1024 images to calculate the respective activation statistics, with MSE averaged over five experiments.}
    \label{fig:mse_aug_vs_non_aug}
\end{minipage}
\end{figure}
\par Training data typically undergoes augmentation before being fed to the model, with BN statistics collected on these augmented images \cite{shorten2019survey}. However, existing data generation methods overlook this step. This is displayed in \cref{eq:bn_orig_stats} and \cref{eq:activation_stats}, where the statistics are calculated on $f_l(x)$ instead of $f_l(\faug(x))$. Our approach directly addresses this by replicating the augmentation process during data generation. 
\par To empirically validate the impact of our augmentation process on the alignment of activation statistics with BN statistics, we performed an experiment comparing MSE across various layers of a ResNet-18 model. This experiment involved running 1024 images, both with and without augmentation, and comparing the resulting activation statistics to the BN statistics of the model. The results, as illustrated in \cref{fig:mse_aug_vs_non_aug}, show that the augmented data consistently exhibited lower MSE values compared to the non-augmented data. These findings support incorporating similar augmentation techniques into the data generation process.

\subsection{Output Distribution Mismatch}\label{sec:output_mismatch}
While most data generation methods for ZSQ focus on minimizing the BNS loss, this approach often fails to match the statistics of feature maps unaffected by this loss. For example, BN layers are absent in the last stages of the network. This mismatch poses a significant challenge to quantization algorithms, as generated data may exhibit mismatched distributions in these feature maps. Consequently, quantization algorithms may choose suboptimal quantization parameters that do not align with the real data distribution, leading to degraded performance. This issue is particularly challenging in hardware-friendly quantization settings, where all layers must be quantized.

\begin{figure}
    \centering
    \centering
    \includegraphics[width=0.6\textwidth]{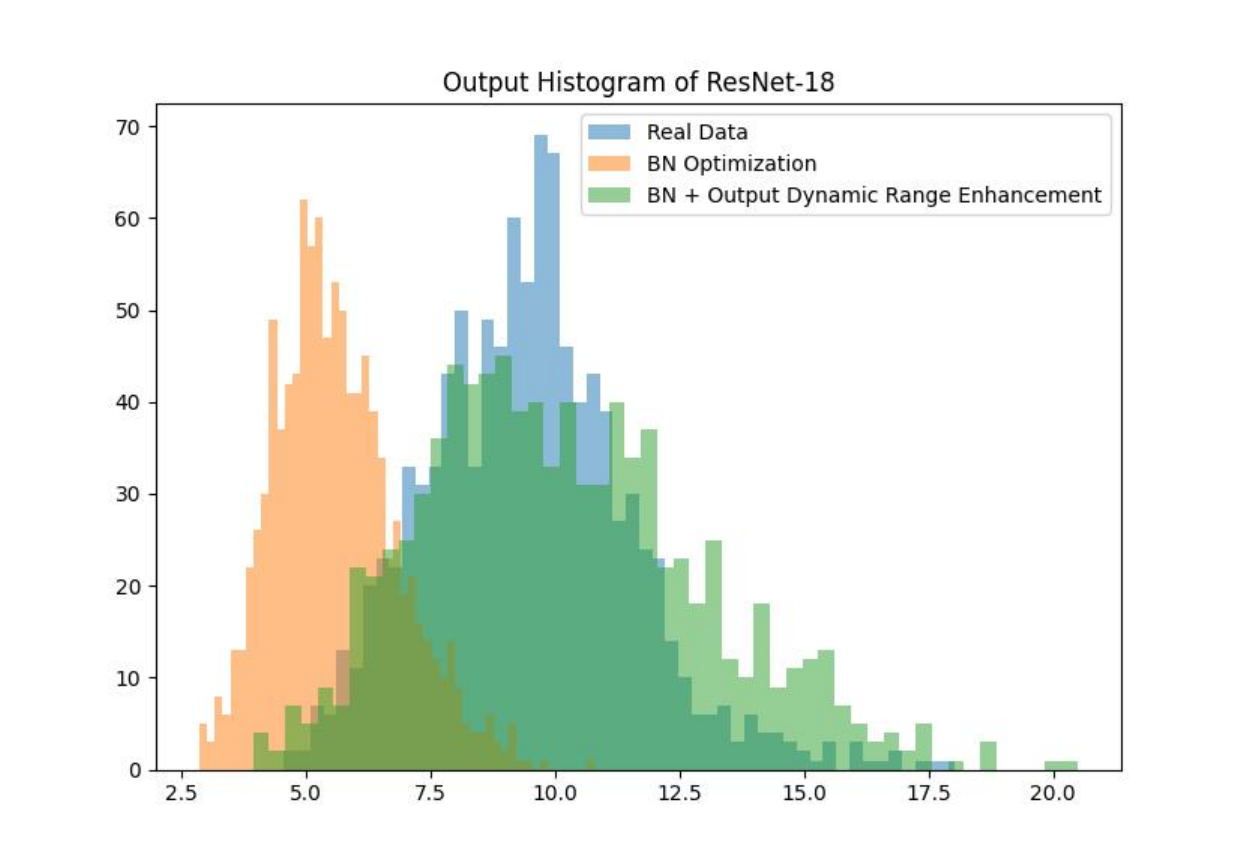}
    \hfill
    \caption{A comparison between the output distributions of a ResNet-18 model when inferring three different data sources: real images, synthetic images generated using only the BNS loss, and synthetic images enhanced with our proposed ODSL. }
    \label{fig:output_enhancment_histogram}
\end{figure}

We illustrate this problem in \cref{fig:output_enhancment_histogram}, which compares the output distributions of ResNet-18 model for real images and generated images using only BNS loss. The figure shows that the logit values produced by the generated images can deviate significantly from those derived from real data. 

Previous methods \cite{Zhong2022_IntraQ, yin2020dreaming,haroush2019,xu2020generative} employed a cross-entropy loss function with a random class to encourage class-dependent image generation. Although this approach partially aligns the support of the distributions by pushing generated images towards strong classification, it has two main limitations: 1) it is confined to classification tasks; 2) it may push the logits to infinity without any constraints, affecting the loss landscape and hindering the convergence of the BNS loss.

To address these limitations, we have introduced an output distribution stretching loss applicable to various tasks beyond classification. Our approach encourages generated images to mimic the output distribution support of real data. 
This strategy assists the quantization algorithm in selecting parameters that more accurately represent the data.

\section{Method}
In this section, we present our proposed method \name{}. It is designed to address the challenges associated with data generation for hardware-friendly ZSQ, where all the model layers, including the output, are quantized. Our method leverages a combination of statistics aggregation to optimize the entire image set simultaneously, image preprocessing to integrate augmentations and image priors, and an output distribution stretching loss that improves the model's output dynamic range. \name{} generates synthetic data that significantly improves performance across various hardware-friendly ZSQ tasks.

\subsection{Statistics Aggregation Scope}
\par We propose a novel approach to aggregate statistics throughout the entire set of generated images, which effectively implies that $N=1$. This approach offers two significant advantages. First, it replicates the process of collecting BN statistics. This leads to better minimization of the BNS loss, as defined in \cref{eq:bn_loss}. 
This is because the statistical estimations computed over a bigger batch become more accurate representations of the dataset statistics.
Second, this approach relaxes the statistical constraints on individual images by considering a larger set of images during optimization. This relaxation allows each image to deviate from the mean and standard deviation of the BN statistics, similar to real data.

\begin{align}
\label{eq:bn_loss_agg}
\mathcal{L}_{BNS}(\dset)= 
\sum_{l=1}^L&\| \mubar(\dset)-\mul\|_2^2 + \| \sigmabar(\dset)-\sigmal \|_2^2. 
\end{align}
\par However, it's important to note that as the set $\dset$ grows in size, computation resources may face limitations in terms of memory capacity, making it impractical to process the entire dataset simultaneously. To address this constraint, we introduce an aggregation algorithm, that effectively estimates the loss in \cref{eq:bn_loss_agg}. This algorithm allows us to optimize the entire set of images and ensure closer alignment of the generated data with the statistics of the pre-trained model's BN layers. We leverage the linearity of expectation and the relationship between variance and second moment to establish the following properties:
\begin{subequations}
\label{eq:agg_stats_activations}
\begin{align}
  &\squareb{\mubar(\dset)}_i 
            = \frac{1}{N}\sum_{n=1}^{N}\squareb{\mubar(\bset_n)}_i\\
    &\squareb{\sigmabar(\dset)}_i = \frac{1}{N}\sum_{n=1}^{N}\squareb{\bar{\vectorsym{M}}^{(2)}_l(\bset_n)}_i - \squareb{\mubar(\dset)}_i^2,\label{eq:agg_stats_activations_var}
\end{align}
\end{subequations}
where
 $\mubar(\bset_n)$, $\squareb{\bar{\vectorsym{M}}^{(2)}_l(\bset_n)}_i=\frac{1}{K d_l}\sum_{k,j=1}^{K,d_l}\squareb{\vectorsym{y}_l^{(k)}}_{i,j}^2$  are the mean and second moment of the $n^{th}$ batch, respectively. The derivation of \cref{eq:agg_stats_activations} is presented in Appendix~\ref{apx:derivation_std_arg}. 
The main idea of the algorithm is to store the statistics $\mubar(\bset_n)$ and $\bar{\vectorsym{M}}^{(2)}_l(\bset_n)$ for all batches. When a batch is optimized, its images are inferred and its layers' statistics are recalculated. Next, the stored statistics from all other batches are averaged together with the new recomputed statistics to recompute the global layers'. Subsequent backpropagation on the loss \cref{eq:bn_loss_agg} is performed only on the specific batch. The detailed steps for computing global statistics are highlighted in red in \cref{alg:data_synthesis}.

This algorithm enables calculating global statistics on any number of images, regardless of memory constraints. By aligning the statistics of the entire image set with the global BN statistics, the generated data more closely matches the statistical properties of the training data.  This, in turn, allows the quantization algorithm to select parameters that better reflect the real data statistics.

\begin{algorithm}
\caption{Image Generation using \name. \textcolor{red}{The global statistics computation is shown in red}, \textcolor{blue}{the image prior computation in blue}, and the \textcolor{MyGreen}{output distribution stretching loss is shown in green.}}
\label{alg:data_synthesis}
\renewcommand{\algorithmicrequire}{\textbf{Input:}}
\renewcommand{\algorithmicensure}{\textbf{Output:}}
\begin{algorithmic}[1]
    \Require A pre-trained model $\mathcal{M}$ with $L$ BN layers.
    \Ensure A set of images: $\dset=\{\x_m\}_{m=1}^M$.
    \State Initialize $\dset$ as standard Gaussian noise and split into $N$ batches.
    \For {each index $n$ in $N$}
        \State \textcolor{blue}{$\tilde{\bset}_n=\{\phi_{prep}(\x_k) :\x_k \in \bset_n \}$}
        \State Compute and store $\{\mubar(\tilde{\bset}_n)\}_{l=1}^{L}, \{\bar{\vectorsym{M}}^{(2)}_l(\tilde{\bset}_n)\}_{l=1}^{L}$.
    \EndFor

    \For{$t \gets 1$ to $T$ iterations}
        \For {each index $n$ in $N$}
            \State \textcolor{blue}{$\tilde{\bset}_n=\{\phi_{prep}(\x_k) :\x_k \in \bset_n \}$}
            \State Run model $\mathcal{M}$ and compute batch statistics $\{\mubar(\tilde{\bset}_n)\}_{l=1}^{L}, \{\bar{\vectorsym{M}}^{(2)}_l(\tilde{\bset}_n)\}_{l=1}^{L}$.
            \State \textcolor{red}{Re-compute the global statistics according to \cref{eq:agg_stats_activations}.}
            \State Compute $\mathcal{L}_{BNS}$ by \cref{eq:bn_loss_agg}.
            \State \textcolor{MyGreen}{Compute $\mathcal{L}_{ODSL}$ by \cref{eq:output_loss}.}
            \State Backpropagate \eqref{eq:total_loss_fn} and update $\bset_n$.
            \State \textcolor{blue}{$\tilde{\bset}_n=\{\phi_{prep}(\x_k) :\x_k \in \bset_n \}$}
            \State \textcolor{red}{Run model $\mathcal{M}$ and compute batch statistics$\{\mubar(\tilde{\bset}_n)\}_{l=1}^{L}, \{\bar{\vectorsym{M}}^{(2)}_l(\tilde{\bset}_n)\}_{l=1}^{L}$.}
        \EndFor 
    \EndFor
\end{algorithmic}
\end{algorithm}

\subsection{Leveraging Data Augmentations and  Image Priors}

\par We introduce a pre-processing pipeline that leverages data augmentations and incorporates image priors to enhance the quality of generated images. We address a previously overlooked aspect regarding the BN statistics where the BN mean and standard deviation saved in the model were derived from an augmented training set. To handle this, we propose incorporating augmentations at the start of each optimization step. Our enhancement strategy contains several carefully crafted preprocessing steps. This step is incorporated using $\phi_{prep}:\mathcal{I}\rightarrow\mathcal{I}$ in \cref{alg:data_synthesis}.

\par We further enhance the process by applying a smoothing filter to add natural properties to the images. This manages two core image generation challenges: 1) preserving the inherent smoothness of natural images without adding complex loss functions like total variation; 2) mitigating checkerboard artifacts, which are crucial for minimizing information loss. Prior studies, such as \cite{46191, jeon2023genie}, have highlighted the importance of these issues. This approach addresses artifact issues similarly to swing-convolutions \cite{jeon2023genie}. However, since the smoothing operation is applied directly to the images, it is not confined to a specific layer, such as convolution. Instead, it mitigates artifact issues across all layers in the model, including those caused by operations like max pooling.

\subsection{Output Distribution Stretching Loss}
To address the output distribution discrepancies described in Section~\ref{sec:output_mismatch}, we propose the output distribution stretching loss (ODSL). This loss enables the generation of data that cover the full support of the model's output and can be applied to any task. The primary objective of the output distribution stretching loss is to maximize the difference between the minimum and maximum values of the model's output for each image. In addition, to prevent significant deviations from the model's typical output values, we impose a constraint per image leveraging the last BN layer. This layer provides a point of information about the activations produced by the model in the training dataset. The objective is given by:
\begin{align}
    \max\limits_{\vectorsym{x}_k} &\|\max\limits_i{\squareb{f_\text{out}(\vectorsym{x}_k)}}_i - \min\limits_i{\squareb{f_\text{out}(\vectorsym{x}_k)}_i}\|_2^2,\\
    \text{s.t.} \quad &\|\bar{\vectorsym{\mu}}^k_L - \vectorsym{\mu}_L\|_2^2 \leq \delta,
    \ \ \ \ \|\bar{\vectorsym{\sigma}}^k_L - \vectorsym{\sigma}_L\|_2^2 \leq \delta, \nonumber
\end{align}
where $\bar{\vectorsym{\mu}}^k_{L}$ and $\bar{\vectorsym{\sigma}}^k_{L}$ represent the mean and standard deviation of the activations at the last BN layer for the $k$-th image. Additionally, $f_\text{out}(\vectorsym{x}_k)$ denotes the output activations of the model for the $k$-th image. Note that $f_\text{out}(\vectorsym{x}_k)$ is a vector, and if the output is not a vector, we flatten it. The parameter $\delta$ defines the allowable deviation, ensuring that the output of each image remains within a controlled range of the typical output statistics of the model. This constraint resembles the alignment of the slack distribution introduced in \cite{Zhang2021_DSG}. The loss function  defined as $\mathcal{L}_{\text{ODSL}}(\dset)=\frac{1}{|\dset|}\sum_{k=1}^{|\dset|} \ell (\vectorsym{x}_k) $:
\begin{align}
\label{eq:output_loss}  
\ell (\vectorsym{x}_k) =   
- \|\max\limits_{i}{\squareb{f_\text{out}(\vectorsym{x}_k)}_i} - \min\limits_{i}{\squareb{f_\text{out}(\vectorsym{x}_k)}_i}\|_2^2    + \max{\left(\|\bar{\vectorsym{\mu}}^k_L - \vectorsym{\mu}_L\|_2^2 - \delta, 0\right)}  + \max{\left(\|\bar{\vectorsym{\sigma}}^k_L - \vectorsym{\sigma}_L\|_2^2 - \delta, 0\right)}.
\end{align}
Subsequently, the ODSL is incorporated into the total loss function (see \cref{alg:data_synthesis}), with the hyperparameter $\lambda$:
\begin{equation}\label{eq:total_loss_fn}
\mathcal{L}(\dset) = \mathcal{L}_{BNS}(\dset) + \lambda\cdot\mathcal{L}_{\text{ODSL}}(\dset).
\end{equation}

\section{Experimental Results}\label{sec:exp}
\begin{table}[t!]
\centering
\caption{  
Top-1 accuracy on ImageNet-1k validation set with models fully quantized using BRECQ \cite{li2021_brecq}. The models are quantized using four different data generation algorithms and real data.}
\label{tab:fully_quantized_results}
\begin{tabular}{cccc} 
\hline
\multirow{2}{*}{Method} & \begin{tabular}[c]{@{}c@{}}ResNet-18\\71.06\end{tabular} & \begin{tabular}[c]{@{}c@{}}ResNet-50\\77.0\end{tabular} & \begin{tabular}[c]{@{}c@{}}MBV2\\72.49\end{tabular}  \\ 
\cline{2-4}
                        & \multicolumn{3}{c}{W4A4 Fully Quantized}                                                                                                                                         \\ 
\hline
\zeroq                  & 0.44 \tstd{0.04}                                                     & 5.47 \tstd{1.66}                                                    & 1.31 \tstd{0.29}                                                        \\
\intraq                  &  \secondbest{47.44}\tstd{2.44}                                                   & 25.6 \tstd{5.58}                                                   & \secondbest{63.45} \tstd{0.26}                                                      \\
\genie                   & 24.86 \tstd{4.81}                                                   & \secondbest{49.84} \tstd{3.55}                                                  & 25.15 \tstd{0.98}                                                      \\
\name{}
(Ours)                    & \best{65.64} \tstd{0.19}                                                    & \best{72.35} \tstd{0.1}                                                   & \best{65.0} \tstd{0.15}                                                       \\
\cdashline{2-4}
Real Data               & 62.5 \tstd{2.01}                                                     & 71.65 \tstd{0.15}                                                 & 60.62 \tstd{0.46}                                                      \\ 
\cline{2-4}
                        & \multicolumn{3}{c}{W2A4 Fully Quantized}                                                                                                                                         \\ 
\cline{2-4}
\zeroq                  & 4.35 \tstd{0.57}                                                    & 21.98\tstd{1.36}                                                   & 0.98\tstd{0.23}                                                        \\
\intraq                  & 8.25 \tstd{2.41}                                                     & 7.41\tstd{0.97}                                                    & 7.97\tstd{1.33}                                                        \\
\genie                   & \secondbest{23.55}\tstd{0.79}                                                    & \secondbest{40.89}\tstd{2.84}                                                   & \secondbest{23.5}\tstd{2.72}                                                        \\
\name{} (Ours)                    & \best{56.55}\tstd{0.64}                                                    & \best{58.87}\tstd{0.29}                                                   & \best{32.9}\tstd{1.14}                                                        \\ 
\cdashline{2-4}
Real Data               & 56.7\tstd{1.71}                                                     & 60.67\tstd{0.47}                                                   & 42.02\tstd{1.52}                                                       \\
\hline
\end{tabular}
\end{table}

This section provides a comprehensive evaluation of \name{}, emphasizing its ability to generate high-quality synthetic data that improves quantization performance across models and PTQ algorithms. We compare several ZSQ algorithms, including \zeroq, \intraq, and \genie$ $, using real data as an additional reference.
Examples of synthetic images generated using \name{} are presented in Appendix \ref{sec:generated_imgs}
\par We begin by comparing the Top-1 accuracy on the ImageNet-1k\cite{ILSVRC15} validation set for ResNets \cite{he2016deep} and MobileNetV2 \cite{sandler2018mobilenetv2} using BRECQ \cite{li2021_brecq} and HPTQ \cite{habi2021hptq} with pre-trained models from \cite{li2024brecq_repo}. The models are quantized in both hardware-friendly and academic quantization settings.

Additionally, we evaluate the proposed method on the COCO validation set for object detection tasks, using models such as RetinaNet \cite{lin2017focal}, SSD \cite{liu2016ssd} and FCOS \cite{tian2020fcos}  from Torchvision \cite{TorchVision_maintainers_and_contributors_TorchVision_PyTorch_s_Computer_2016}. Finally, we conduct an ablation study to show the contribution of each component of \name{}.
In all tables, the best results are shaded in gray and highlighted in bold (\best{x}), and the second-best results are highlighted in bold (\secondbest{x}).
The implementation details for our experiments can be found in Appendix \ref{sec:impl_details}.

\subsection{Classification Results}
Here, we examine the effectiveness of image generation for hardware-friendly quantization, where all model layers are quantized, including the output. First, we employ the BRECQ algorithm \cite{li2021_brecq} tailored for this setting. We quantize all weights and activations to W$w$A$a$, where $w$ is the number of bits for the weights and $a$ for the activations. This differs from the original BRECQ, which assigns 8 bits to specific layers and activations.

In \cref{tab:fully_quantized_results}, we present the Top-1 accuracy on the ImageNet-1k validation set using  BRECQ algorithm \cite{li2021_brecq} with activation bit-width set to 4 bits and weights bit-width is set to 2 and 4 bits. These results demonstrate that \name{} significantly outperforms all other approaches and is comparable to, and occasionally surpasses, the results obtained using real data.

Second, we employ the HPTQ method \cite{habi2021hptq}, which focuses on quantization of efficient\footnote{By efficient we mean the use of power-of-two thresholds.} hardware platforms. The quantization bit-width for all layers is set to 8-bit for both weights and activations, ensuring compatibility with widely used hardware accelerators. In \cref{tab:real_world_deployment_results}, we present the Top-1 accuracy on the ImageNet-1k validation set using the HPTQ quantization method. These results show a smaller performance gap due to the less challenging 8-bit configuration.  We further analyze the effect of varying bit widths, validating this observation, with detailed results provided in Section \ref{sec:varying_bit_widths} of the supplementation materials. Additionally, \intraq $ $ demonstrates substantial improvement benefiting from its CE-like loss function.  Overall, as seen in \cref{tab:fully_quantized_results} and \cref{tab:real_world_deployment_results}, these results highlight the advantages of \name{}, achieving superior quantization performance on real-world deployment scenarios. 
\begin{table}[t!]
\centering
\caption{
Top-1 accuracy on ImageNet-1k validation set with models fully quantized to W8A8 using using hardware-friendly PTQ \cite{habi2021hptq}. The models are quantized using four different data generation algorithms and real data.}
\label{tab:real_world_deployment_results}
\begin{tabular}{cccc} 
\hline
Method    & \begin{tabular}[c]{@{}c@{}}ResNet-18\\71.06\end{tabular} & \begin{tabular}[c]{@{}c@{}}ResNet-50\\77.0\end{tabular} & \begin{tabular}[c]{@{}c@{}}MBV2\\72.49\end{tabular}  \\ 
\hline
\zeroq     & 2.21\tstd{0.00}                                                     & 9.34\tstd{0.02}                                                    & 1.79\tstd{0.01}                                                        \\
\intraq    & \secondbest{70.60}\tstd{0.05}                                                     & \secondbest{75.95}\tstd{0.02}                                                   & \secondbest{72.05}\tstd{0.03}                                                       \\
\genie     & {64.21}\tstd{0.10}                                                    & 74.18\tstd{0.03}                                                   & 65.70\tstd{0.04}                                                        \\
\name{} (Ours)      & \best{70.91}\tstd{0.02}                                                    & \best{76.54}\tstd{0.03}                                                    & \best{72.47}\tstd{0.03}                                                       \\
\cdashline{1-4}
Real Data & 70.91\tstd{0.03}                                                    & 76.58\tstd{0.01}                                                   & 72.46\tstd{0.03}                                                       \\
\hline
\end{tabular}
\end{table}

In addition to the hardware-friendly quantization scheme, we compare \name{} to \zeroq{}, \intraq{}, and \genie{} on the academic quantization scheme where \name{} has shown competitive results. The results are shown in Appendix \ref{sec:academic_quant_res}.

\subsection{Object Detection Results}
Here, we present the results of applying \name{} to object detection models, showcasing its versatility across tasks beyond classification. We selected three widely used object detection architectures for our experiments: RetinaNet \cite{lin2017focal}, SSD \cite{liu2016ssd} and FCOS \cite{tian2020fcos}  from Torchvision \cite{TorchVision_maintainers_and_contributors_TorchVision_PyTorch_s_Computer_2016}. The models were fully quantized using AdaRound \cite{nagel2020_adaround}, with 4-bit weights and 8-bit activations, and data generated using \name{}, \genie, and \zeroq. Since \intraq $ $ requires a CE loss, it was not suitable for this task. We evaluated the quantized models on the COCO validation dataset \cite{lin2014microsoft} and present the mean average precision (mAP). As shown in \cref{tab:object_detection_results}, \name{} significantly outperforms all other methods and matches the accuracy achieved with real data.

\begin{table}
\centering
\caption{mAP on the COCO validation dataset using AdaRound with four different data generation algorithms and real data.}
\label{tab:object_detection_results}
\begin{tabular}{cccc} 
\hline
Method    & \begin{tabular}[c]{@{}c@{}}SSDLite\\MBV3\\Large\\~21.3\end{tabular} & \begin{tabular}[c]{@{}c@{}}FCOS\\ResNet50\\FPN V1~\\39.2\end{tabular} & \begin{tabular}[c]{@{}c@{}}RetinaNet\\ResNet50 \\FPN V2\\~41.5\end{tabular}  \\ 
\hline
\zeroq     & \secondbest{14.58}\tstd{0.0}                                                                       & \secondbest{37.54}\tstd{0.0}                                                                   & 2.60\tstd{0.01}                                                                        \\
\genie     & 13.71\tstd{0.01}                                                                       & 28.08\tstd{0.01}                                                                   & \secondbest{12.89}\tstd{0.02}                                                                        \\
\name{} (Ours)      & \best{17.20}\tstd{0.01}                                                                        & \best{37.67}\tstd{0.0}                                                                   & \best{32.28}\tstd{0.01}                                                                        \\
\cdashline{1-4}
Real Data & 17.16\tstd{0.01}                                                                       & 37.98\tstd{0.0}                                                                   & 32.36\tstd{0.0}                                                                        \\
\hline
\end{tabular}
\end{table}

\subsection{Ablation Study of \name{}}
In these experiments, we conduct an ablation study to validate the necessity of each component of \name{}, as presented in \cref{tab:ablation_study}. We systematically modify individual components to assess their impact on quantization accuracy, using the hardware-friendly quantization setting described in the previous subsection. The results demonstrate that while each component individually contributes to improvements over the baseline, the most significant gains arise when combining distribution alignment and ODSL.
\par
\begin{figure}[ht]
    \centering
        \centering
        \includegraphics[width=0.8\textwidth]{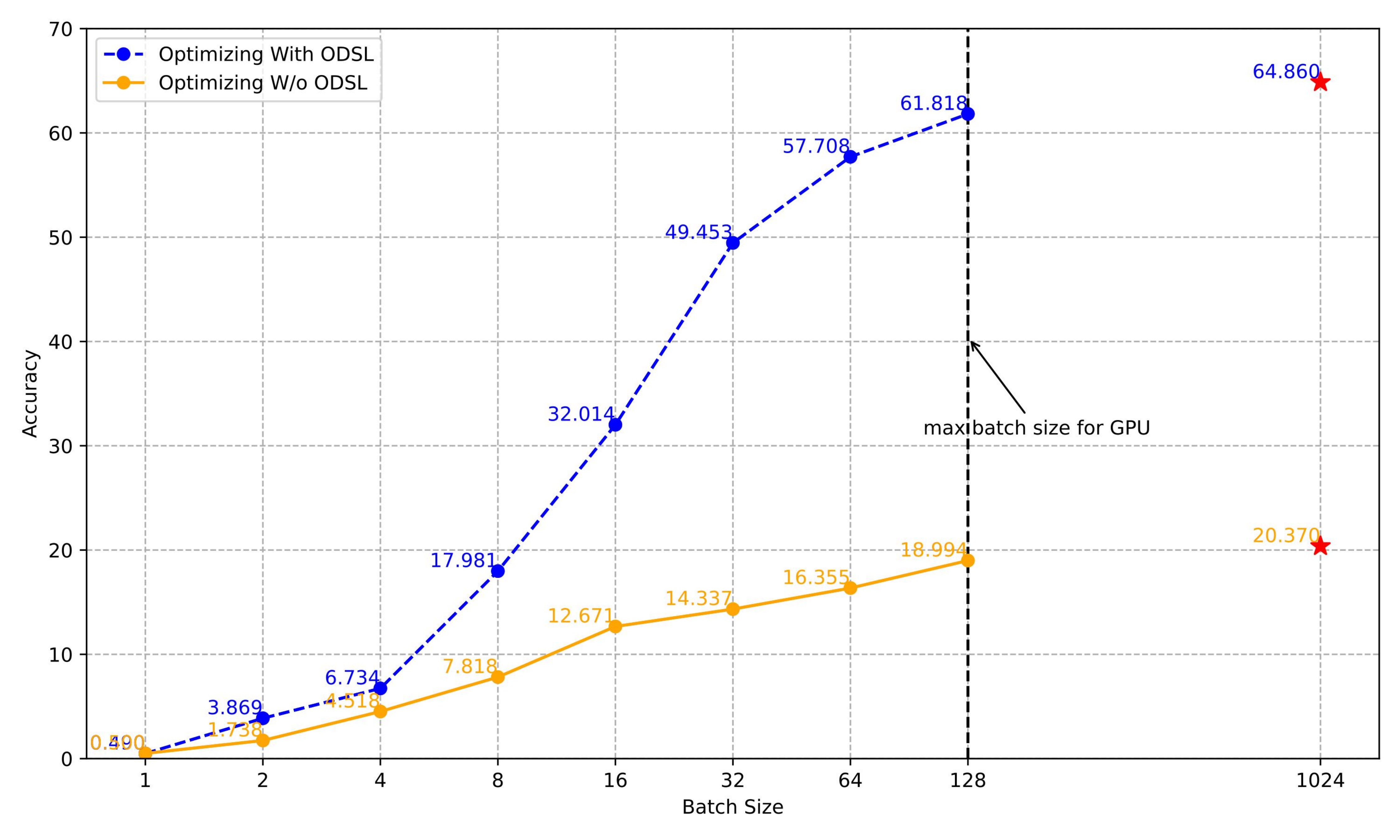}
    \caption{The figure represents the Top-1 accuracy on the ImageNet-1k validation set of ResNet-18 quantized to W4A4-bit precision using AdaRound with 1024 generated images. The blue curve represents optimizing with ODSL, while the orange curve represents optimizing without ODSL.
    The x-axis denotes the statistics aggregation scope (batch size) used in the image generation process, where each batch is optimized separately according to \cref{eq:bn_loss}. The red stars indicate the results of using \name{}'s aggregation algorithm.
    }
    \label{fig:odsl_batch_size_effect}
\end{figure}

 To further investigate this effect, we conducted an additional experiment to assess how quantization performance benefits from ODSL and increasing statistical aggregation scope. We performed ZSQ using images generated with and without ODSL, varying the aggregation scope sizes, and evaluated the resulting accuracy. As shown in \cref{fig:odsl_batch_size_effect}, the advantage of ODSL becomes evident with as few as four images, and the performance gap widens as the optimization scope expands. A larger statistical aggregation scope provides an additional degree of freedom for each individual image to deviate from the global statistics of the image set.
\begin{table}
\centering

\caption{Ablation study evaluating the impact of \name{}'s components on the ImageNet-1k validation set Top-1 accuracy using the BRECQ quantization algorithm applied to ResNet18 and MobileNet-V2. The study focuses on analyzing the effects of three key components: Statistics Aggregation Scope (SAS), Image Preprocessing (IP), and Output Distribution Stretching Loss (ODSL).}
\label{tab:ablation_study}

\begin{tabular}{ccccccc} 
\hline
\multirow{2}{*}{SAS}  & \multirow{2}{*}{IP}     & \multirow{2}{*}{ODSL}     & \multicolumn{2}{c}{ResNet18}                           & \multicolumn{2}{c}{MBV2}                               \\
                     &                         &                         & W4A4                       & W2A4                      & W4A4                      & W2A4                       \\ 
\hline
                     &                         &                         & 15.92                      & 15.4                      & 18.53                     & 20.79                      \\
\cmark                  &                         &                         & 20.37                      & 24.73                     & 24.58                     & 20.61                      \\
                     & \cmark                     &                         & 17.0                       & 18.0                      & 18.97                     & 24.02                      \\
                     &                         & \cmark                     & 25.15                      & 24.42                     & 31.15                     & 23.42                      \\
\cmark                  & \cmark                     &                         & 22.58                      & 22.79                     & 24.91                     & 24.83                      \\
\cmark                  &                         & \cmark                     & \secondbest{64.86}                      & \secondbest{50.74}                     & \secondbest{62.44}                     & \secondbest{25.46}                      \\
 & \cmark & \cmark & 19.54 & 23.51 & 26.47 & 25.43  \\
\cmark                  & \cmark                     & \cmark                     & \best{65.64}             & \best{56.55}            & \best{65.0}             & \best{32.90}              \\
\hline
\end{tabular}
\end{table}

\section{Conclusion}
In this paper, we present \name, a novel data generation method designed for hardware-friendly ZSQ. Our approach addresses three key limitations of existing methods: the statistics calculation scope of generated images, not accounting for data augmentations in the generation process, and feature maps missing BN comparison points. To resolve the statistics alignment issue, we introduce an aggregation algorithm that compares the statistics of the entire image set to the BN statistics during each iteration. For data augmentation, we add a preprocessing stage that applies image augmentations and a smoothing filter at the start of each iteration. Lastly, to address the missing optimization points, we introduce an output distribution stretching loss, applicable to any task. The experimental results demonstrate that \name{} achieves state-of-the-art results for hardware-friendly ZSQ, where the models are fully quantized, enhancing the practicality of deploying quantized models on resource-constrained hardware. Additionally, the versatility of our method is validated through its successful application to various tasks, including image classification and object detection, and quantization algorithms such as BRECQ, AdaRound, and HPTQ, highlighting its robustness and efficiency in real-world scenarios. Future research will explore the extension of \name{} to transformer models, investigating the potential benefits and optimizations that \name{} can bring to the quantization of these architectures.

\bibliographystyle{splncs04}
\bibliography{ref}
\newpage
\appendix
\section{Appendix}
\subsection{Implementation Details}
\label{sec:impl_details}

For all experiments, unless stated otherwise, we use the following hyperparameters: for classification tasks, 1024 images are generated, and for object detection, 32 images are used. Each result represents an average of five experiments. We perform 1000 optimization iterations, using the RAdam optimizer \cite{liu2019variance} with an initial learning rate of 16 and a reduced learning rate on the plateau scheduler.
\par For $\phi_{prep}$, we begin by applying a $3\times3$ Gaussian smoothing filter. Then, we randomly apply horizontal flipping to the images with a probability of 0.5. Next, we perform random cropping by starting with images that are 32 pixels larger in both height and width than the final output size and then cropping them to the desired output shape. The final images are center-cropped.

\subsection{Derivation of Equation \eqref{eq:agg_stats_activations}}\label{apx:derivation_std_arg}
Here, we provide the derivation of \cref{eq:agg_stats_activations}. We begin with the data mean decomposition, which is followed by the decomposition of variance. 

\begin{align}
    \squareb{\mubar(\dset)}_i &\triangleq \frac{1}{M}\sum_{m=1}^{M}{\squareb{f_l(\x_m)}_i}\nonumber\\
    &=\frac{1}{N}\sum_{n=1}^{N}\frac{1}{K}\sum_{\x_k \in \bset_n}{\squareb{f_l(\x_k)}_i}\nonumber\\
            &= \frac{1}{N}\sum_{n=1}^{N}\squareb{\mubar(\bset_n)}_i
\end{align}

where $\mathcal{S}_n$ is the set of index correspond to the $n^{th}$ batch.
The first step follows directly from the definition in \eqref{eq:activation_stats_mean}. In the second step, we apply the linearity of summation. Finally, we use the notion of an empirical mean over batch. Next, we present the relationship between the variance and second moment:

\begin{align}
    \squareb{\sigmabar(\dset)}_i &\triangleq\frac{1}{M}\sum_{m=1}^{M}\brackets{\squareb{f_l\brackets{\x_m}}_i - \squareb{\mubar(\dset)}_i}^2\nonumber\\
    &=\squareb{\mubar(\dset)}_i^2+\frac{1}{M}\sum_{m=1}^{M}\squareb{f_l\brackets{\x_m}}_i^2 -2\cdot \squareb{f_l\brackets{\x_m}}_i\squareb{\mubar(\dset)}_i  \nonumber\\
    &=\frac{1}{N}\sum_{n=1}^{N}\frac{1}{K}\sum_{\x_k\in\bset_n}\squareb{f_l\brackets{\x_k}}_i^2 -\squareb{\mubar(\dset)}_i^2  \nonumber\\
    &= \frac{1}{N}\sum_{n=1}^{N}\squareb{\bar{\vectorsym{M}}^{(2)}_{\ell}(\bset_n)}_i - \squareb{\mubar(\dset)}_i^2,
\end{align}

The first step follows from the definition in \eqref{eq:activation_stats_var}. In the second step, we expand the parentheses. In the third step, we apply the linearity of summation and the notion of empirical mean. Finally, we use the notion of an empirical moment over the batch.

\subsection{Additional Results}\label{sec:additional_res}

\subsubsection{Academic Quantization Results} \label{sec:academic_quant_res}
We validate our approach within the academic quantization scheme. In this setup, the first and last layers and the input to the second layer are quantized to 8 bits, while the output layer remains unquantized. \cref{tab:academic_quantization_results} shows the Top-1 accuracy on the ImageNet-1k validation set using the BRECQ quantization algorithm under the academic quantization scheme, with activation bit-width set to 4 bits and weight bit-width set to 2 and 4 bits. The results demonstrate that the \name{} achieves competitive results to \genie $ $, while greatly improving the image generation runtime since \name{} does not use a generator. Specifically, \genie $ $ requires approximately two and a half hours on a V100 GPU to generate 1024 images for ResNet18, while \name{} takes less than half an hour on an RTX 3090. Although \zeroq $ $ and \intraq $ $ may offer faster generation times than \name{}, they deliver inferior performance in both academic and hardware-friendly quantization schemes.

below the caption
\begin{table}
\centering
\caption{Top-1 accuracy on ImageNet-1k validation set with models quantized using BRECQ \cite{li2021_brecq} in an academic quantization setting. The models are quantized using four different data generation algorithms and real data}.
\label{tab:academic_quantization_results}
\begin{tabular}{cccc} 
\hline
\multirow{2}{*}{Method} & \begin{tabular}[c]{@{}c@{}}ResNet-18\\71.06\end{tabular} & \begin{tabular}[c]{@{}c@{}}ResNet-50\\77.0\end{tabular} & \begin{tabular}[c]{@{}c@{}}MBV2\\72.49\end{tabular}  \\ 
\cline{2-4}
                        & \multicolumn{3}{c}{W4A4 Academic Quantized}                                                                                                                                      \\ 
\hline
\zeroq                   & 68.67\tstd{0.09}                                                    & 74.26\tstd{0.12}                                                   & \best{69.27}\tstd{0.22}                                                       \\
\intraq                  & 67.78\tstd{0.27}                                                    & 66.08\tstd{0.67}                                                   & 69.06\tstd{0.13}                                                       \\
\genie                   & \best{69.67}\tstd{0.05}                                                    & \best{74.90}\tstd{0.06}                                                    & 69.13\tstd{0.05}                                                       \\
\name{} (Ours)                    & \secondbest{69.56}\tstd{0.04}                                                    & \secondbest{74.72}\tstd{0.12}                                                   & \secondbest{69.23}\tstd{0.08}                                                       \\
\cdashline{2-4}
Real Data               & 69.74\tstd{0.07}                                                    & 74.90\tstd{0.06}                                                    & 69.30\tstd{0.10}                                                        \\ 
\cline{2-4}
                        & \multicolumn{3}{c}{W2A4 Academic Quantized}                                                                                                                                      \\ 
\cline{2-4}
\zeroq                   & 59.73\tstd{0.17}                                                    & 63.53\tstd{0.22}                                                   & 27.42\tstd{1.04}                                                       \\
\intraq                  & 49.17\tstd{0.56}                                                    & 44.03\tstd{1.76}                                                   & 22.8\tstd{3.17}                                                        \\
\genie                   & \best{64.37}\tstd{0.17}                                                    & \best{69.53}\tstd{0.04}                                                   & \best{48.23}\tstd{3.24}                                                       \\
\name{} (Ours)                    & \secondbest{64.15}\tstd{0.10}                                                    & \secondbest{68.91}\tstd{0.06}                                                   & \secondbest{45.83}\tstd{1.44}                                                       \\
\cdashline{2-4}
Real Data               & 65.86\tstd{0.10}                                                    & 70.28\tstd{0.03}                                                   & 54.46\tstd{1.44}                                                       \\
\hline
\end{tabular}
\end{table}

\subsubsection{The Effect of Varying Bit Widths }\label{sec:varying_bit_widths}
\begin{figure}[ht]
    \centering
        \centering
        \includegraphics[width=0.6\textwidth]{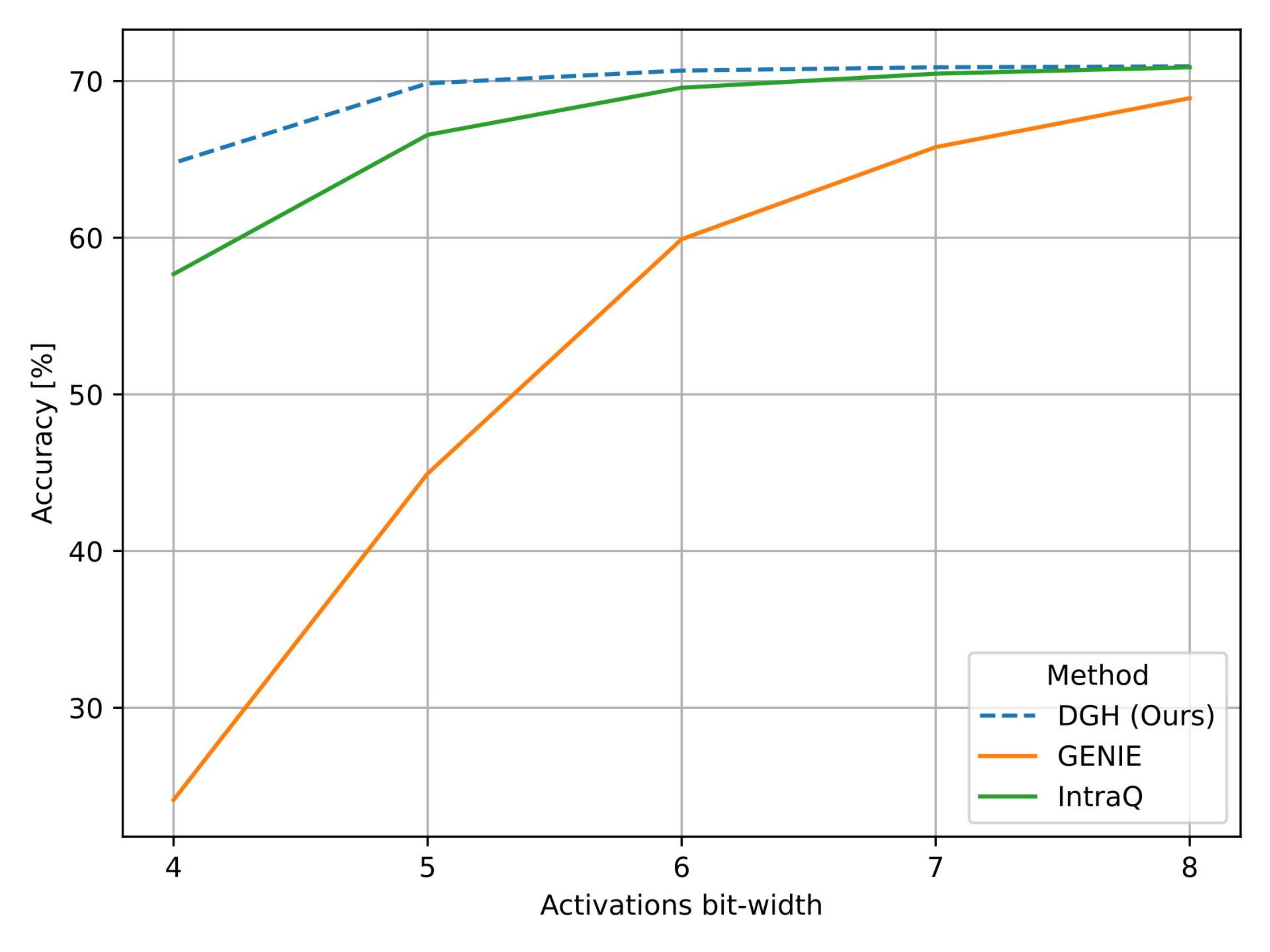}
    \caption{Top-1 accuracy on the ImageNet-1k validation set using the BRECQ quantization algorithm in a hardware-friendly quantization setting, with various activation bit-widths while the weight bit-width is fixed at 8 bits. The y-axis represents the accuracy, and the x-axis represents the activation bit-width.}    \label{fig:act_sweep}
\end{figure}

\begin{figure}[ht]
    \centering
        \centering
        \includegraphics[width=0.6\textwidth]{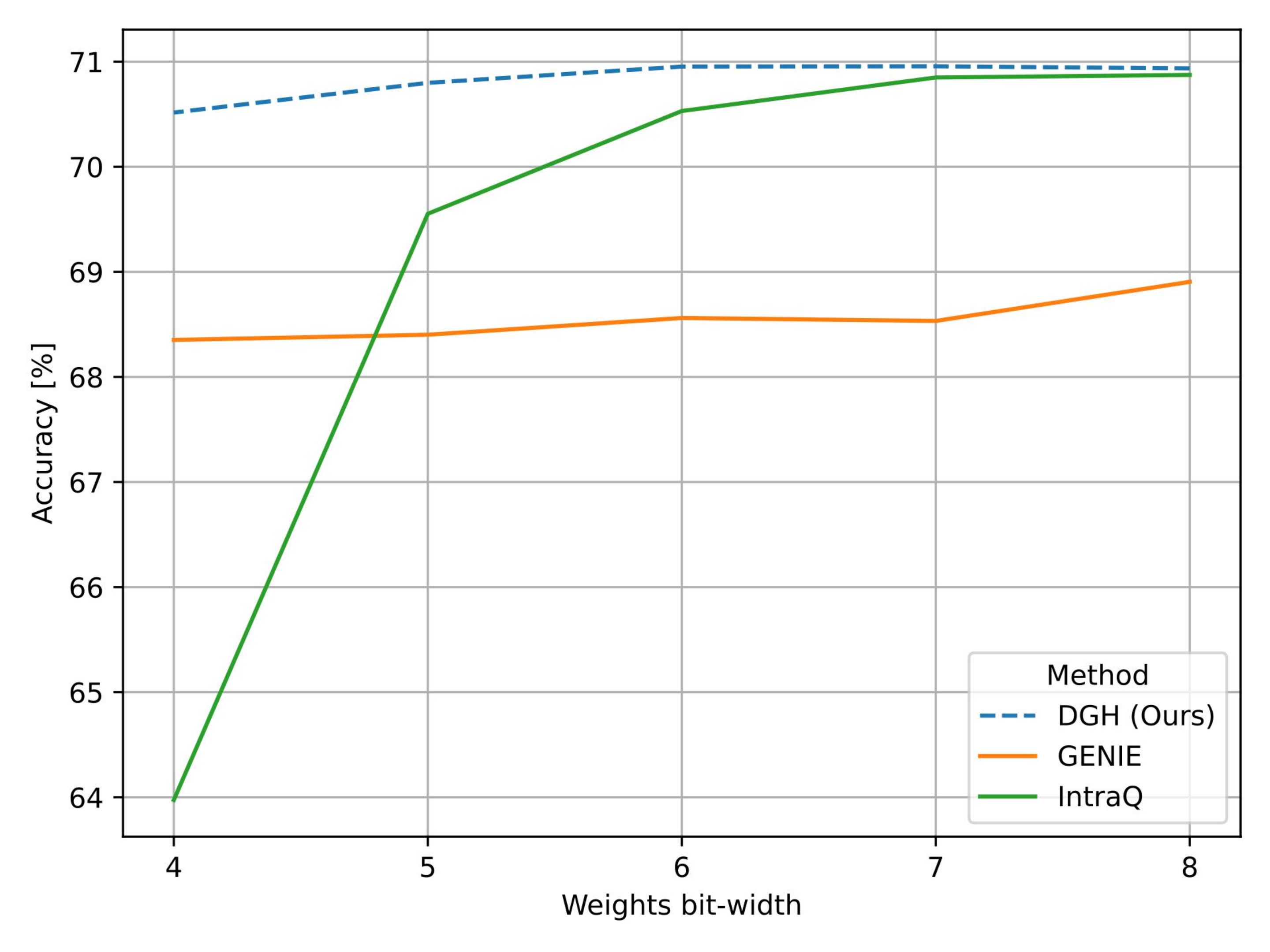}
    \caption{Top-1 accuracy on the ImageNet-1k validation set using the BRECQ quantization algorithm in a hardware-friendly quantization setting, with various weight bit-widths while the activation bit-width is fixed at 8 bits. The y-axis represents the accuracy, and the x-axis represents the weights bit-width.}
\label{fig:w_sweep}
\end{figure}

Here, we provide results offering further insights into the performance of \name. We present the Top-1 accuracy on ImageNet-1k validation set using the BRECQ quantization algorithm in the hardware-friendly quantization setting, tested across various bit-widths. In this experiment, we keep the weight bit-width fixed at 8 bits and evaluate \intraq{}, \genie{}, and \name{} at different activation bit-widths. The results are shown in \cref{fig:act_sweep}.  Next, we fix the activation bit-width at 8 bits and evaluate \intraq{}, \genie{}, and \name{} at varying weight bit-widths, with the results presented in \cref{fig:w_sweep}. From both sets of results, we observe that the performance gap between \name{} and \intraq{}, \genie{} increases as the bit-width decreases. Note that we omit \zeroq{} as its accuracy is consistently near zero in all cases.

\subsubsection{Generated Images Using \name{} }\label{sec:generated_imgs}

\begin{figure}[ht]
    \centering
    \includegraphics[width=\textwidth]{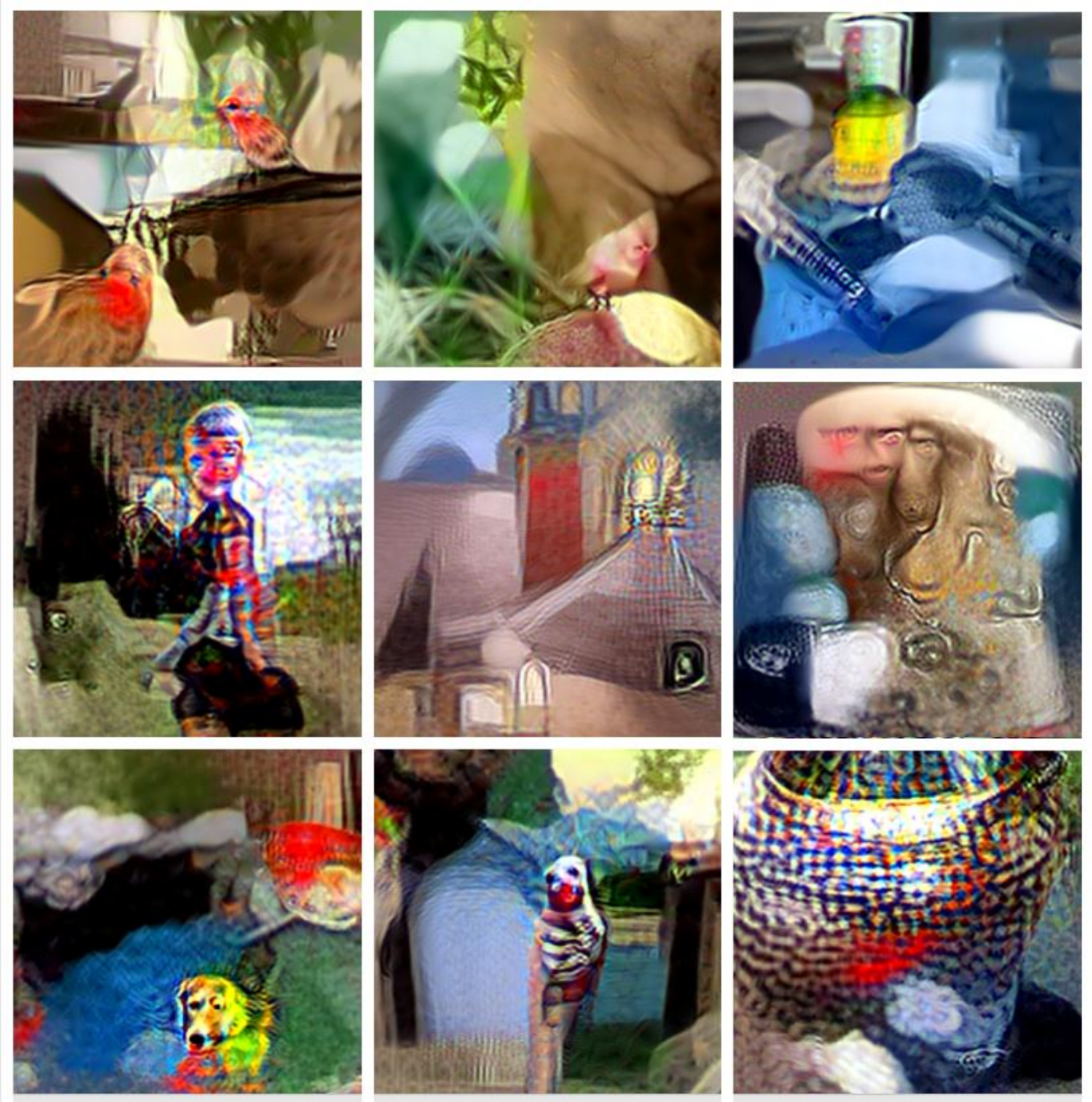}
    \caption{Images generated from a ResNet18 using \name{} using 1k Iterations.}
    \label{fig:generated_images}
\end{figure}

\begin{figure}[ht]
    \centering
    \includegraphics[width=\textwidth]{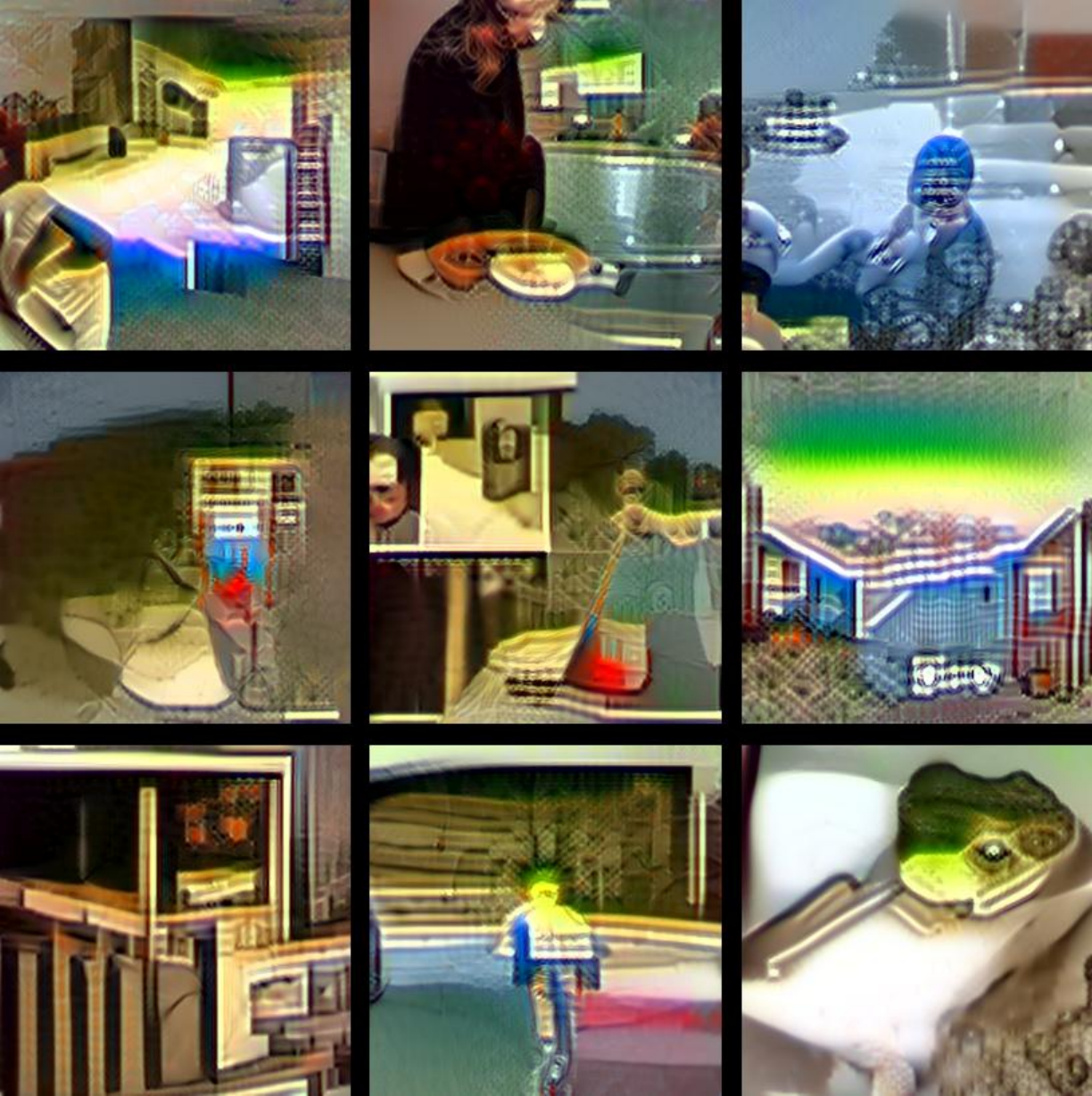}
    \caption{Images generated from a ResNet18 using \name{} using 40k Iterations.}
    \label{fig:generated_images_40k}
\end{figure}

Next, we visualized the images generated by \name{} from ResNet18 after 1k iterations, as shown in (\cref{fig:generated_images}), which were used in all experiments Additionally, we present images generated after 40k iterations in (\cref{fig:generated_images_40k}). In \cref{fig:generated_images}, we observe the initial emergence of shapes and patterns in the generated images. As shown in \cref{fig:generated_images_40k}, with additional iterations, these patterns and class-specific features become increasingly distinct, reflecting improved image quality and clearer representation of different classes in the generated images.

\end{document}